\documentclass[conference]{IEEEtran}

\usepackage[utf8]{inputenc}
\usepackage{amsmath}
\usepackage{amssymb}
\usepackage{algpseudocode}
\usepackage{algorithm}
\usepackage{hyperref}       
\usepackage{graphicx}
\usepackage[export]{adjustbox}
\usepackage{subcaption}
\usepackage{cite}
\usepackage{array}
\usepackage{amsthm}

\newcolumntype{C}[1]{>{\centering\arraybackslash}p{#1}}
\begin{document}
\title{LanFL: Differentially Private Federated Learning with Large Language Models using Synthetic Samples}

\author{\IEEEauthorblockN{1\textsuperscript{st} Huiyu Wu}
\IEEEauthorblockA{\textit{Industrial Engineering and Management Sciences} \\
\textit{Northwestern University}\\
Evanston, USA \\
huiyuwu2025@u.northwestern.edu}
\and
\IEEEauthorblockN{2\textsuperscript{nd} Diego Klabjan}
\IEEEauthorblockA{\textit{Industrial Engineering and Management Sciences} \\
\textit{Northwestern University}\\
Evanston, USA \\
d-klabjan@northwestern.edu}
}


\date{}
\newtheorem{assumption}{Assumption}
\newtheorem{theorem}{Theorem}
\newtheorem{proposition}{Proposition}
\newtheorem{lemma}{Lemma}
\newtheorem{corollary}{Corollary}
\newtheorem{mechanism}{Mechanism}

\maketitle

\begin{abstract}
Federated Learning (FL) is a collaborative, privacy-preserving machine learning framework that enables multiple participants to train a single global model. However, the recent advent of powerful Large Language Models (LLMs) with tens to hundreds of billions of parameters makes the naive application of traditional FL methods to LLMs impractical due to high computational and communication costs. Furthermore, end users of LLMs often lack access to full architectures and weights of the models, making it impossible for participants to fine-tune these models directly. This paper introduces a novel FL scheme for LLMs, named LanFL, which is purely prompt-based and treats the underlying LLMs as black boxes. We have developed a differentially private synthetic sample generation mechanism to facilitate knowledge sharing among participants, along with a prompt optimization scheme that enables learning from synthetic samples. Our extensive experiments demonstrate that LanFL successfully facilitates learning among participants while preserving the privacy of local datasets across various tasks.
    
\end{abstract}

\section{Introduction}
Federated Learning (FL) is a privacy preserving machine learning scheme that allows many participants to train a single global model without sharing their local training samples \cite{fedavg}. FL relies on participants sharing their local gradient updates or models, and a central server or some decentralized framework to aggregate the updates \cite{fl_survey, bcfl_survey}. Since only gradient updates or models are shared, each client's local data set privacy is preserved. This is especially crucial nowadays as there are increasing regulations on data sharing such as EU's General Data Protection Regulation \cite{gdpr} and the California Consumer Privacy Act \cite{CA_privacy}. It has been applied in many fields including finance, edge computing, and healthcare. However, despite its popularity, FL still faces challenges in privacy threats, heterogeneity, and communication overhead, and new paradigms for FL are needed \cite{fl_challenges_app, sabfl}.

Large Language Models (LLMs) are sophisticated deep neural networks based on transformer architectures, designed to comprehend, generate, and manipulate natural language \cite{attention}, demonstrating performances that approach, and sometimes exceed, human capabilities, LLMs excel in tasks such as logical reasoning, translation, and complex examinations \cite{gpt4}. Their exceptional abilities render them applicable across various domains similar to those in Federated Learning (FL), including customer service, finance, law, and education \cite{comprehensive_overview}. Despite their widespread adoption and impressive functionalities, LLMs are associated with significant safety concerns, including risks of fraud, impersonation, and misinformation dissemination \cite{misuse}. Consequently, developers restrict access to the models' weights, architecture, and parameter details for prominent LLMs like Chat-GPT 4, Claude 3, and Gemini \cite{gpt4, claude3, gemini}. It is also worth noting that with these models containing tens or hundreds of billions of parameters, transmitting complete models several times or conducting extensive fine-tuning is often impractical. Instead, prompt engineering emerges as a viable alternative to leverage these models effectively \cite{prompt_survey}. Prompts serve as inputs to LLMs; it has been demonstrated that the outputs from these models are highly sensitive to the nature of the prompts provided. Well-constructed prompts enhance the relevance, logical coherence, and accuracy of the outputs.

At a first glance, the application of FL with LLMs appears impractical. Federated Learning primarily relies on local gradient updates or local models; however, in the context of LLMs, users often lack access to the model weights. Furthermore, even if FL clients could access the weights of LLMs, computing gradients over numerous iterations would not be computationally efficient. Additionally, the significant communication overhead associated with transmitting potentially hundreds of billions of parameters renders the process inefficient. Nevertheless, there are some methods available to enable FL for LLMs, which typically involve introducing trainable parameters into the LLM architecture while maintaining the pre-trained components unchanged \cite{fl-llm, fl-llm-adapter}. It is important to recognize that although these existing LLM FL methods address the issues of large communication and training costs, they still necessitate knowledge of the specific architecture and parameters of the underlying LLM. This requirement restricts the practical applications of such approaches. For instance, a consortium of medium-sized hospitals is likely not to prefer to invest in open source LLMs and going down a rabbit hole to buy costly GPUs and hire experts. Instead, they may opt to utilize the best pre-trained LLMs available from large technology companies, where the architectural details and parameters remain confidential. 

There are also several notable distinctions between FL with LLMs and traditional FL approaches. Traditionally, FL typically involves training a model from scratch, whereas FL with LLMs generally starts with pre-trained LLMs, focusing on fine-tuning these models for specific tasks rather than training new ones from the ground up. Another key difference lies in model ownership and accessibility: in conventional FL, clients own the models and have full access to all model aspects. However, this is not always the case with FL involving LLMs. For instance, banks participating in an FL scheme using LLMs might opt to utilize powerful pre-trained models from third-party providers without the intention or need to train their own models, thereby lacking access to the model's architecture and weights. Additionally, unlike traditional FL where often clients use the same models, in the LLM context, different participants may choose to use various third-party models based on factors such as cost or other specific requirements. 

Inspired by the capability of LLMs to learn from contextual examples in prompts and their ability to generate synthetic samples \cite{fewshot, tabular_syn}, we propose a new LLM FL scheme, LanFL, which aims to bridge the gap between FL and LLMs. LanFL is designed to enable participants to engage in FL without requiring access to the underlying architecture and weights of the LLM. On a high level, in one LanFL step, participants first create synthetic samples using their local LLMs. Then the participants generate prompts using the synthetic samples and their own knowledge. The next step is participants sharing the prompts to other participants. Since LLMs can learn from prompts, combining local data and the prompts with synthetic data received, participants can improve model performance without accessing other participants' local data sets. We also show that LanFL is differentially private. In the experiments LanFL performs well when clients have heterogeneous data sets. 

Our contributions are as follows. First, to the best of our knowledge, we are the first work enabling FL for LLMs using only prompts, without accessing and modifying underlying weights and architectures. This advancement is crucial as it circumvents the need for participants to access the weights directly, thereby also allowing for the use of different underlying LLMs by various participants. Second, we introduce an innovative method for generating and selecting synthetic samples that ensures their effectiveness while also confirming that they are sufficiently distinct from the original training samples. Additionally, our mechanism for synthetic sample generation is designed to be differentially private. Third, our experimental results indicate that LanFL performs robustly across a variety of datasets, and different levels of data heterogeneity.

This paper is organized as follows. In Section 2 we examine related works, in Section 3 we explain the details of LanFL operations and study the differential privacy properties of LanFL, in Section 4 we explore our comprehensive experiment results, and finally in Section 5 we present the conclusions.

\section{Related Works}
There are a few works focusing on FL of LLMs, and all requires full knowledge of the architecture of the underlying model. Che et al propose FedPepTAO \cite{fl-llm} which utilizes techniques in prefix-tuning and adaptive optimization. On a high level, participants perform prefix-tuning by adding trainable parameters to each layer of the LLM \cite{prefix}. In each round a set of participants only update prompt parameters of specific layers. FedPepTAO is parameter efficient since each round only a few layers of prefixes are updated, and it does not modify the weights of the underlying LLM. However, it still requires full knowledge of the underlying LLM architecture and does not scale well as future LLMs increase in the number of layers and parameters. A concurrent work by Hou et al, PrE-Text \cite{pretext}, employs differentially private synthetic samples to fine-tune LLMs. Their approach generates synthetic samples through a multi-round process using only textual data, and the fine-tuning step requires knowledge of the underlying model architecture. In contrast, our approach generates synthetic samples in a single round and accommodates numerical features. Additionally, we leverage in-context learning techniques, which eliminate the need for knowledge of the underlying model architecture.

Similarly, Kim et al propose using adapter mechanisms to LLMs for FL \cite{fl-llm-adapter}. Specifically, they add adapter layers such as LoRA \cite{lora} into the LLM architectures, and only fine-tune the parameters of added layers while keeping the pre-trained LLM parameters frozen. Fan et al implement this framework named FATE-LLM that utilizes trainable adapters \cite{fate}. Again, this approach requires knowing the exact architecture of the underlying LLM and all experiments are run with not the larges LLMs having less than 10 billion parameters. 

Another idea for FL LLM is proposed by Su et al \cite{titanic}. The TITANIC framework they propose first splits the LLM into multiple parts by layers, and each participant receive a part of the LLM. FL is facilitated by having each participant perform a forward pass, sending intermediate outputs to the next participant who possesses the subsequent segment, followed by a backward loss propagation step. While TITANIC demonstrates superior performance compared to previous frameworks, it is not without challenges; specifically, it encounters issues related to privacy and significant communication overhead. In contrast, our LanFL approach is based on the exchange of prompts not requiring weights, which circumvents these issues, offering a more efficient and secure method of collaborative learning.

Prompting techniques are also closely related to our work. They involve using specific inputs to elicit high-quality and accurate responses from LLMs. The effectiveness of these outputs is heavily dependent on the construction of well-designed prompts \cite{prompt_survey}. Few-shot prompting \cite{fewshot} supplies LLMs with example inputs that guide the model to produce outputs that are not only formatted like the examples but also contextually relevant to the topics addressed by these examples. Chain-of-thought (CoT) prompting, another related technique, encourages LLMs to articulate a sequence of logical reasoning steps before arriving at a final answer. This method has been shown to enhance the logical coherence of the outputs and is particularly effective in logic reasoning tasks \cite{cot}. Building on the CoT framework, approaches such as self-consistency \cite{consistency_prompt} and tree-of-thought \cite{tree-ot} recognize that different reasoning pathways can lead to correct answers. These methods allow LLMs to generate multiple reasoned responses before selecting the most appropriate final output, thus enhancing the robustness and reliability of the model’s decision-making process.

The generation of synthetic datasets using LLMs represents a key advancement upon which our LanFL relies. Frameworks such as ZeroGen, ProGen, and SuperGen facilitate production of synthetic datasets by conditioning on labels and employing task-specific models to iteratively select and filter synthetic samples \cite{zerogen}, \cite{progen}, \cite{supergen}. These processes operate independently of human-annotated samples and treat pre-trained LLMs as black boxes. Nonetheless, these frameworks primarily focus on textual data, whereas LanFL also accommodates numerical data. Furthermore, in FL operations, participants typically possess well-annotated samples, a feature that LanFL leverages. ClinGen represents another LLM-based framework for generating synthetic samples, specifically targeting medical tasks; however, it does not generalize to numerical data \cite{clingen}. Borisov et al introduce GReaT, a framework designed for generating tabular data \cite{tabular_syn}. We build upon this framework by incorporating advanced prompting techniques to enhance its efficacy with numerical tabular data.

\section{LanFL}
\subsection{LanFL Operations}
\begin{figure}[t]
  \centering
  \includegraphics[height=5.6cm]{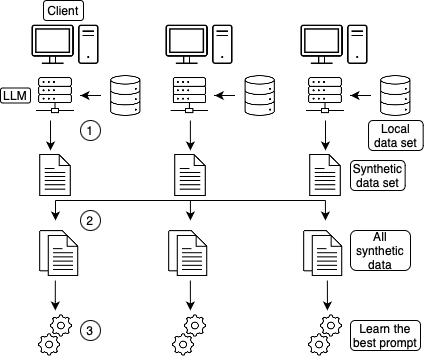}
  \caption{LanFL Operations. Step 1: Clients generate synthetic data sets by prompting LLMs using local data sets. Step 2: Clients share the synthetic data sets among themselves. Step 3: Clients learn the best prompt utilizing the synthetic data sets received.}
  \label{Lanfl_png}
\end{figure}

As discussed in the introduction, FL with LLMs are challenging because of the large model sizes and black-box nature of the models. Our proposed LanFL method addresses these challenges and is the first to employ a prompt-based framework for FL with LLMs. It enables clients to engage in FL while treating the underlying LLMs as black boxes. As shown in Figure \ref{Lanfl_png}, at a high level, LanFL consists of three steps: generate synthetic samples, share knowledge, and learn using prompts. LanFL steps start with each participant creating synthetic samples and generate logic solutions to the synthetic samples based on its local training samples. The next step is to share the synthetic samples with labels and reasoning among all other participants. Finally, participants add all synthetic samples to the local training set, and learn from the samples by optimizing for the best prompt and use the optimized prompt for downstream tasks.

In detail, the initial step involves participants generating synthetic samples using LLMs with few-shot prompting. Specifically, participants randomly select a few local samples that include chain-of-thought reasoning and use these as examples to prompt the LLM to generate new samples. This process is repeated to create a synthetic dataset. The chain-of-thought reasoning for the local samples can be either manually obtained or generated by prompting the LLM as a separate pre-processing step. Consequently, the resulting synthetic samples contain not only features and labels but also the chain-of-thought reasoning, making them immediately usable as examples in prompts. The second step involves the sharing of knowledge, wherein participants exchange the synthetic data sets they have generated. Since these synthetic data sets encompass chain-of-thought reasoning derived from the participants' local samples, the act of sharing these data sets inherently facilitates the dissemination of knowledge. Finally, upon receiving synthetic datasets from other participants, the clients aim to learn from these samples by designing few-shot prompts that optimize in-context learning. The learning is relied on LLMs' in-context learning capabilities \cite{incontext_survey}, and by utilizing the synthetic samples containing knowledge from other participants, privacy is achieved.

\subsection{Synthetic Samples}\label{synthetic_section}
A key element in the LanFL framework is the generation of synthetic samples. Inspired by chain-of-thought and few-shot prompting \cite{cot, fewshot}, we propose the following two-step mechanism to generate synthetic samples. For a participant with local data set $D=\{(x_1, y_1), (x_2, y_2), \dots, (x_n, y_n)\}$, and an LLM $L$ with temperature set to $0$, the first step is to generate the chain of thought reasoning $r_i$ for each of the sample $(x_i, y_i)$. Specifically 
\begin{equation*}
    r_i = L(M_r(x_i, y_i)).
\end{equation*}
Here $M_r$ can be any function that converts $(x_i, y_i)$ into prompt so that the LLM outputs a chain-of-thought reasoning $r_i$ for specific sample $(x_i, y_i)$. Note that $x_i, y_i, r_i$ and the output of function $M_r$ are all sequences of tokens. The first step can be skipped if the participant has readily available reasoning steps in the local data set or it decides to manually provide chain-of-thought reasoning for all the local samples. During the second step, the participant randomly selects $k$ samples, $D_k=\{(x_{s_1}, r_{s_1}, y_{s_1}), (x_{s_2}, r_{s_2}, y_{s_2}), \dots, (x_{s_k}, r_{s_k}, y_{s_k})\}$, and uses these samples as examples in the prompt to generate a synthetic sample
\begin{equation*}
    syn_{D_k} = L(M_{syn}(D_k)).
\end{equation*}
We can then extract $(x_{D_k}, r_{D_k}, y_{D_k})$ from $syn_{D_k}$. Here $M_{syn}$ is a prompt template that converts the samples into a few-shot prompt that instructs the LLM to generate a synthetic sample. To create the synthetic data set, the participant simply repeats the previous step to reach the desired number of synthetic samples.

\begin{figure}[t]
  \centering
  \includegraphics[height=7cm]{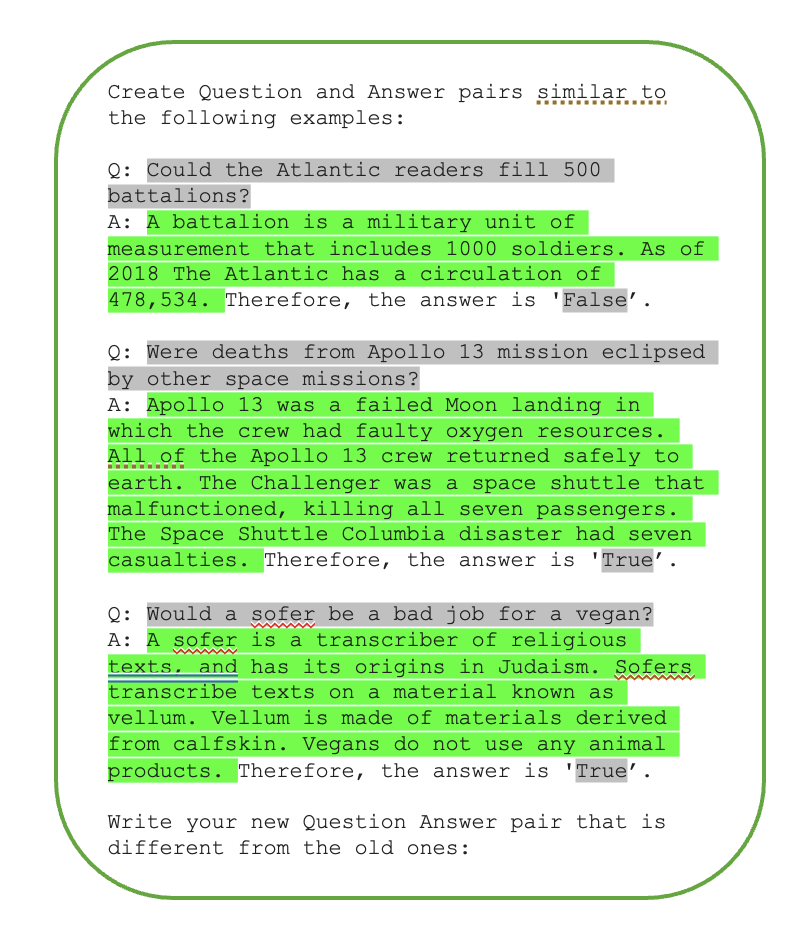}
  \caption{Example prompt used to generate synthetic sample (Output of $M_{syn}$)}
  \label{prompt_png}
\end{figure}

To be more concrete, $M_r$ can simply list the tokens $x_i$, and asks LLM why $y_i$ is the answer. For example, the output of $M_r$ can be: \emph{`Weng earns \$12 an hour for babysitting. Yesterday, she just did 50 minutes of babysitting. How much did she earn? Think step by step and answer why she earned \$10.'} For questions that require domain knowledge to generate logic reasoning, the participant can first manually provide logical reasoning for a few selected samples, add background knowledge, and use them as examples in prompt and instruct the LLM to provide reasoning for other samples. For synthetic sample generation, Figure \ref{prompt_png} shows an example output of $M_{syn}$ using the CommonsenseQA data set \cite{csqa}. The sentences marked gray are $x_i$ and $y_i$, sentences marked green are chain-of-thought reasoning, and the remaining sentences are meta prompt and formatting added by $M_{syn}$. It is important to note that by incorporating logical reasoning and facts into the examples within the prompt, the resulting synthetic samples encompass not only the questions and their answers but also the associated logical reasoning steps. This approach is more efficient, as it allows for the simultaneous generation of synthetic samples and their logical reasoning, rather than producing the synthetic samples first and subsequently generating the reasoning as a separate step.

For numerical features, we need one pre-processing step to convert the features into tokens that LLMs can understand better. Instead of the simple textual encoder in GReaT where we add `\emph{is}' between feature name and the feature value\cite{tabular_syn}, we require the participant to connect the features into a cohesive paragraph. It may require domain knowledge for specific tasks. As an example, for features concerning a credit card holder's information, instead of using `\emph{Gender is male, education is high school, age is 30, payment is 400, usage is 600,}' we use `\emph{A high-school educated 30 year-old male has used \$600 of his credit and has made a \$400 payment.}' as $x_i$.

Another consideration for a participant is that of all the synthetic samples created, the participant is recommended to discard samples that are similar to those in its local data set. For example, for each of the synthetic sample $x_{syn}$, the participant should compute $n$ BLEU scores $b_1, b_2, \dots b_n$ for each of the local sample $x_1, x_2, \dots x_n$ \cite{bleu}. Then the participant can set a threshold $t$ depending on the application and discard any synthetic samples that have $b_{max}>t$ where $b_{max}=max(b_1, b_2, \dots b_n)$. For numerical features, the participant can utilize the $L2$ distance between the feature vectors to determine whether or not to discard any of the synthetic samples.

\subsection{Differential Privacy}
Differential Privacy is a formal definition of privacy and it characterizes if an output mechanism leaks information if the underlying data set differs by one sample \cite{dp, dpml, dp_results}. To establish our synthetic sample generation method is differential privacy, we first formalize the synthetic sample generation mechanism.

\begin{mechanism}\label{mechanism}
    We define the random synthetic sample generation mechanism $f$ as follows. Let us have two numbers $N, k\in \mathbb{N}$, and $N\geq k+1$. The mechanism $f$ generates synthetic sample $f(D)=x_{syn}$ given a data set $D\in dom(f)$ by following the following two steps. First, we select $A\subseteq D$ such that $|A|=k$ by uniform randomly selecting elements one by one from $D$ without replacement. Then, we apply function $L$ to $A$ to generate $L(A)=x_{syn}$, where $L(\cdot)$ is any deterministic function that outputs a sequence of tokens. We also restrict the $dom(f)$ to satisfy $D\in dom(f)$, $|D|\geq N$.
\end{mechanism} 

A technicality is that here the function $L$ applies to the set $A$, thus we are implicitly assuming the ordering of the samples in $A$ does not matter. The results are similar if ordering of the elements in A matters (epsilon and delta take different values, but the proof is similar). Note that $L$ is deterministic and the only randomness of $f$ comes from the sampling of the subset $A$. The restriction on the domain of $f$ means we can only apply the mechanism if we have a data set that has more than $N$ samples. In the FL with LLM, we can pick $N$ to be the smallest number of samples of the data sets among the FL participants. 

Note that the mechanism $f$ is the same as the procedure to generate one synthetic sample in LanFL. Specifically, in LanFL the set $A$ is the data set containing the $k$ random local examples, and becuase we set the temperature of the LLM to $0$, $M_{syn}$ and the LLM is deterministic function $L$. Now we introduce the differential privacy result.

\begin{theorem}\label{theorem1}
    Mechanism \ref{mechanism} is $(\delta, \epsilon)$ differentially private for $\delta=\frac{k}{N+1}$ and $\epsilon=0$. Specifically, for any two data sets $D_1, D_2$ in the domain of $f$ that differ by only one element by addition or deletion, and any subset $S$ in the range of $f$ we have
    \begin{equation*}
        P(f(D_1)\in S)\leq \frac{k}{N+1}+e^{0}P(f(D_2)\in S).
    \end{equation*}
\end{theorem}

Theorem \ref{theorem1} shows that our synthetic sample generation mechanism for a single synthetic sample is $(\delta, \epsilon)$ differentially private by definition. Generally we have $N \gg k$ meaning our sample size is much greater than the number of samples in the prompt used to generate synthetic samples, therefore $\delta$ is small indicating a strong privacy guarantee. The proof of Theorem \ref{theorem1} is in the Appendix. In practice, participants generate several synthetic samples, and we have the following Corollary showing generating $w$ synthetic samples is also differentially private.
\begin{corollary}\label{corollary}
    We call $f_w$ the process of performing Mechanism \ref{mechanism} for a total of $w$ times generating $w$ synthetic samples. Then Mechanism $f_w$ is $(\delta, \epsilon)$ differentially private for $\delta=\frac{kw}{N+1}$ and $\epsilon=0$. Specifically, for any two data sets $D_1, D_2$ in the domain of $f_w$ that differ by only one element by addition or deletion, and any subset $S$ in the range of $f_w$ we have
    \begin{equation*}
        P(f(D_1)\in S)\leq \frac{kw}{N+1}+e^{0}P(f(D_2)\in S).
    \end{equation*} 
\end{corollary}
This Corollary follows from the Sequential Theorem in \cite{dpml}. It is important to note that, in practical applications, we generate only $w$ synthetic samples, which usually represents a small fraction of the overall size of the dataset $N$. Therefore, $\frac{kw}{N+1}$ is small, indicating a strong differential privacy guarantee.

\subsection{Prompt Optimization}
The last step in the LanFL operations is participants optimizing for a best prompt utilizing all the synthetic samples received from other participants in addition to their local data sets. It is known that LLMs are sensitive to input prompts, for example, the number and ordering of the examples in the prompt matters \cite{sensitivity, bayesian, ordering, incontext_survey}. Therefore, by optimizing the number of samples and portions of synthetic samples used as examples in a prompt, participants should learn from other participants' samples.

It is impractical to iterate over all the combinations of samples and synthetic samples in the prompt. Therefore, we propose Algorithm \ref{optimization}, a greedy optimization method. First, the participant sets aside a portion of the local data set as a test set, and let the remaining samples be a training set. Then the participant selects a different number of samples from the training set as examples in the prompt and evaluates the results on the test set, and finds the optimal number of samples to include in the prompt (lines $2$-$10$). After collecting all the synthetic samples and using the best number of samples $n_1$ in the prompt, the participant finds the best number of synthetic samples $n_2$ to use in prompts by testing on the test set again (lines $12$-$21$). Algorithm \ref{optimization} summarizes the operations to obtain the number of total samples and the number of synthetic samples in the prompt. The cost can be interpreted as the number of incorrect answers or the mean squared error for continuously valued tasks.

Another benefit of the LanFL optimization is by comparing the cost of using the optimal number of training samples $c_{train}$ and the cost of using the optimal number of synthetic samples with training samples $c_{syn}$, we can examine the effectiveness of LanFL operations. If $c_{syn}-c_{train}>0$, then we know it is beneficial to receive the synthetic samples from other participants and LanFL achieves the goal of learning. Finally, we want to point out that loops in lines $4$ and $15$ should be repeated a few times to reduce the randomness in results.

\begin{algorithm}[t]
\caption{LanFL Optimization}\label{optimization}
\begin{algorithmic}[1]
\State Initialize training set $X_{train}$ and test set $X_{test}$
\For{$n = 1, 2, 3, \dots$ }
    \State Set cost $c_n=0$
    \For{Each test sample $x_{t}\in X_{test}$} 
        \State Randomly select $x_{i_1}, x_{i_2} \dots x_{i_n} \in X_{train}$
        \State \parbox[t]{\dimexpr0.9\linewidth-\algorithmicindent}{Compute cost $c$ using some metric between $Y_{train}$ and $Y_{predicted}$}
        \State Update cost $c_n \mathrel{+}= c$
    \EndFor
\EndFor
\State Record optimal cost $c_{train} = min\{c_i\}$, $1\leq i\leq n$
\State Optimal number of samples in prompt $n_1=argmin_{i}\{c_i\}$, $1\leq i\leq n$
\State Collect synthetic samples $X_{syn}$
\For{$n = 1, 2, 3, \dots n_1$}
    \State Set cost $c_n=0$
    \For{Each test sample $x_{t}\in X_{test}$} 
        \State Randomly select $x_{i_1}, x_{i_2} \dots x_{i_{n}} \in X_{syn}$
        \State Randomly select $x_{i_n}, x_{i_{n+1}} \dots x_{i_{n_1}} \in X_{train}$
        \State \parbox[t]{\dimexpr0.9\linewidth-\algorithmicindent}{Compute cost $c$ using some metric between $Y_{train}$ and $Y_{predicted}$}
        \State Update cost $c_n \mathrel{+}= c$
    \EndFor
\EndFor
\State Record optimal cost $c_{syn} = min\{c_i\}$, $1\leq i\leq n_1$
\State Optimal synthetic samples in prompt $n_2=argmin_{i}\{c_i\}$, $1\leq i\leq n_1$
\end{algorithmic}
\end{algorithm}

\section{Experiments}
\subsection{Synthetic Samples Evaluation}
We evaluate our synthetic samples using two criteria. First, how different are they from the training samples, and second, how good are they in downstream tasks. For experiments in this section, we use PaLM 2 as the underlying LLM and Strategy Question and Answering (StrategyQA) as our data set \cite{palm2, strategyqa}. We perform the synthetic sample generation mechanism with $k=3$.

\begin{figure}[t]
  \centering
  \includegraphics[height=5.6cm]{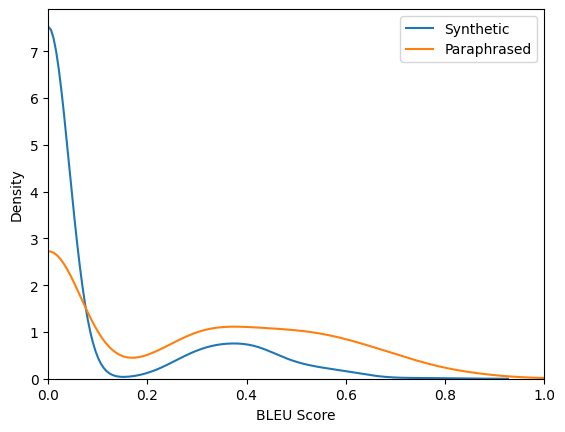}
  \caption{BLEU scores distributions for synthetic samples and paraphrased training samples}
  \label{syn_bleu_scores}
\end{figure}

Although we have already showed differential privacy of our synthetic sample generation mechanism, the property only states that probabilistically one cannot infer whether one specific sample is in the training data set based on the synthetic sample. However, there is no guarantee that synthetic samples are different from the training samples. To show our mechanism produces sufficiently different samples, we computed $b_{max}$ for each our synthetic samples in section \ref{synthetic_section} \cite{bleu}. Additionally, we also paraphrased the training samples used to generate synthetic samples using the same LLM and computed $b_{max}$ for each of the paraphrased sample as a benchmark. Figure \ref{syn_bleu_scores} shows the distribution of $b_{max}$ of synthetic and paraphrased samples. We find not only the synthetic samples are different from the original samples by having small BLEU scores, but also significantly different from paraphrased training samples as well. 

To quantify how good the synthetic samples are, we propose the following experiments. We test the performance of LLM on the same task using the following $4$ different prompts, zero-shot no prompt, three-shot using training set samples with CoT reasoning, three-shot using synthetic samples with CoT reasoning, and three-shot with cohesive but irrelevant paragraphs. The results are summarizes in Table \ref{syn_diff_res}. We find that using synthetics samples as examples in the prompt show a similar performance to using training samples, and they are both better than not using examples or using irrelevant examples. Since the goal of LanFL is to use the synthetic samples to represent the knowledge in the training samples, a similar performance indicate that synthetic samples are indeed good for LanFL operations.

\begin{table}[t!]
\centering
\caption{Synthetic Samples Results (\%)} \label{syn_diff_res}
\begin{tabular}{ c|C{0.75cm}|C{0.75cm}|C{0.75cm}|C{0.75cm} } 
  & 0-shot &  3-shot train & 3-shot synth & 3-shot other \\ \hline
 StrategyQA & 69.75 & 74.67 & 74.76 & 70.20 \\
\end{tabular}
\end{table} 

\subsection{LanFL Experimental Setup}
The goal of the experiments is to show LanFL operations enables learning, therefore in the following experiments we aim to show after receiving synthetic samples from other clients, a participant can obtain better results by adjusting prompts compared to using only its local data set. For each of the experiments, we first aside $10\%$ samples as the validation set and $10\%$ of samples as the test set, then partition the data set into $10$ subsets representing the participants and each of the participants use the local samples to generate $100$ synthetic samples. We aggregate the synthetic samples together and perform Algorithm \ref{optimization}, LanFL optimization, on the validation set to select the best $n_1$ and $n_2$, and finally, we test the performance on the test set using the optimal mixture of synthetic samples. We use Gemini-1.5-Flash \cite{gemini}, Llama3.1-70B \cite{llama3}, and Mixtral-8$\times$7B \cite{mistral} in our experiments.

\subsection{LanFL Results}
The experiment is conducted on the UCI Credit Card Default dataset \cite{credit_card}, a structured dataset consisting of features such as past credit card payments, education, and gender. The objective is to predict whether an individual will default on payments the following month. We follow the LanFL procedure, and the experimental results are summarized in Table \ref{results}. We compare the F1 score of models on the test set using LanFL, after receiving all synthetic samples, to the results from in-context learning using only original samples. The F1 score is chosen as our metric due to the dataset's high imbalance, and our greater interest in identifying defaults. Additionally, we include a comparison to a benchmark weighted random guess. From the results, we observe that all tested LLMs show improved performance after incorporating synthetic samples in the prompt using LanFL optimization. Furthermore, the results significantly outperform the random guess benchmark with LanFL.

\begin{table}[t!]
\centering
\caption{Experimental Results (F1 score \%)} \label{results}
\begin{tabular}{ c|C{0.75cm}|C{0.75cm}|C{0.75cm}|C{0.75cm}} 
  & Gemini &  Mixtral & Llama & Random Guess\\ \hline
 UCI-LanFL & 33.11 & 37.96 & 36.30 & 22.12 \\
 UCI & 8.40 & 13.79 & 30.30 & 22.12 \\
\end{tabular}
\end{table}


\section{Conclusion}
In this study, we introduced LanFL, a prompt-based FL scheme specifically designed for LLMs. LanFL offers the advantage of treating models as black boxes, enhancing its applicability in a wide range of real-world scenarios. This scheme integrates breakthroughs in synthetic sample generation with LLMs and sophisticated prompt engineering techniques. As an FL scheme, LanFL safeguards participant privacy through our differentially private synthetic sample generation mechanism and facilitates learning via optimized prompting strategy.Our experiments demonstrate that the proposed LanFL scheme effectively facilitates learning from all participants. Notably, utilizing synthetic samples from other participants enhances test set performance. The results are also robust across various popular LLMs.

\bibliographystyle{IEEEtran}
\bibliography{IEEEabrv,citations}

\section*{Appendix}
\subsection*{A. Proof of Theorem \ref{theorem1}}
\emph{Proof.} Consider two data sets $D_1, D_2$ that differ by only one element by addition/deletion in the domain of mechanism $f$. Without loss of generality, assume $D_1=D_2\cup \{x_a\}$ for some element $x_a$ and $|D_1|=c$. We have $|D_2|=c-1$. 

For any subset $S$ in the range of mechanism $f$, let $n_1$ be the number of subset $E\subseteq D_2$ such that $|E|=k$ and $L(E)\in S$, and $n_2$ be the number of subsets $E\subseteq D_1$ such that $|E|=k, x_a\in E$ and $L(E)\in S$. Here $n_1$ and $n_2$ may or may not be independent, but individually, their ranges are
\begin{equation*}
    0\leq n_1 \leq {c-1 \choose k},\hspace{0.5cm} 0\leq n_2 \leq {c-1 \choose k-1}.
\end{equation*}

We then compute 
\begin{equation*}
    P(f(D_1)\in S)=\frac{n_1+n_2}{{c \choose k}},\hspace{0.5cm} P(f(D_2)\in S)=\frac{n_1}{{c-1 \choose k}}.
\end{equation*}

Taking the difference we have
\begin{equation}\label{diff1}
\begin{split}
    P(f(D_1)\in S)-P(f(D_2)\in S)&=\frac{\frac{{c-1 \choose k}}{{c \choose k}}n_1+\frac{{c-1 \choose k}}{{c \choose k}}n_2-n_1}{{c-1 \choose k}}\\
    &=\frac{\frac{c-k}{c}n_2-\frac{k}{c}n_1}{{c-1 \choose k}}\\
    &\leq \frac{\frac{c-k}{c}{c-1 \choose k-1}-\frac{k}{c}*0}{{c-1 \choose k}}\\
    &=\frac{k}{c}.
\end{split}
\end{equation}

Additionally, we also have
\begin{equation}\label{diff2}
\begin{split}
    &\hspace{0.5cm}P(f(D_2)\in S)-P(f(D_1)\in S) \\
    &= \frac{n_1}{{c-1 \choose k}} - \frac{n_1+n_2}{{c \choose k}}\\
    &\leq \frac{n_1}{{c-1 \choose k}} - \frac{n_1}{{c \choose k}}\\
    &=\frac{n_1 - \frac{c-k}{c}n_1}{{c-1 \choose k}}\\
    &=\frac{\frac{k}{c}n_1}{{c-1 \choose k}}\\
    &\leq \frac{k}{c}.
\end{split}
\end{equation}

Note that inequality (\ref{diff1}) implies that $P(f(D_1)\in S)\leq \frac{k}{c}+e^0P(f(D_2)\in S)$, and inequality (\ref{diff2}) implies that $P(f(D_2)\in S)\leq \frac{k}{c}+e^{0}P(f(D_1)\in S)$. Therefore, for any two data sets $D_1, D_2$ that differ by only one element by addition/deletion in the domain of mechanism $f$, and for any subset $S$ in the range of mechanism $f$, we have
\begin{equation*}
    P(f(D_1)\in S)\leq \frac{k}{m}+e^{0}P(f(D_2)\in S),
\end{equation*}
where $m=min(|D_1|,|D_2|)+1$. By the restriction placed on $dom(f)$, $min(|D_1|,|D_2|)\geq N$, we have $m \geq N+1$. Note that $\frac{k}{m}$ decreases in $m$ for $0<k<N+1$, therefore,
    \begin{equation*}
        P(f(D_1)\in S)\leq \frac{k}{N+1}+e^{0}P(f(D_2)\in S).
    \end{equation*}\null\hfill $\blacksquare$\\
\end{document}